# Highly Accurate Prediction of Jobs Runtime Classes

Anat Reiner-Benaim
Anna Grabarnick

Department of Statistics
University of Haifa
Haifa, Israel
areiner@stat.haifa.ac.il
ann@gmail.com

Edi Shmueli

Intel Corporation
Haifa, Israel
edi.shmueli@intel.com

*Abstract*— separating the short jobs from the long is a known technique to improve scheduling performance. In this paper we describe a method we developed for accurately predicting the runtimes classes of the jobs to enable this separation. Our method uses the fact that the runtimes can be represented as a mixture of overlapping Gaussian distributions, in order to train a CART classifier to provide the prediction. The threshold that separates the short jobs from the long jobs is determined during the evaluation of the classifier to maximize prediction accuracy. Our results indicate overall accuracy of 90% for the data set used in our study, with sensitivity and specificity both above 90%.

## I. INTRODUCTION

Supplying job schedulers with information on how long the jobs are expected to run enabled the development of the backfilling algorithms, which leverage this information to pack the jobs more efficiently and improve system utilization [1]. These algorithms, however, were designed for parallel systems, in which the jobs require many processors in order to execute, and processor fragmentation (idleness) is a big concern. In those environments the scheduler needs to know the actual runtimes of the jobs (use numeric predictions) to be able to optimize the schedule and improve performance [10].

Our work targets systems in which most jobs are serial, like server farms that are used for software testing. In those environments sophisticated scheduling algorithms are not required, and in order to improve performance it is enough to simply separate the short jobs from the long and assign them to different queues in the system [12]. This separation reduces the likelihood that short jobs will be delayed after long ones, improves the average turn-around times of the jobs and overall system throughput. (Figure 1)

Respectively, to implement such a system it is enough to only predict the runtime classes of the jobs – whether they will be short or long, in order to assign them to the right queue. On the other hand, any misclassification of the jobs can severely impact performance. For example, mistakenly assigning long jobs to the short jobs queue will cause many of the short jobs to be delayed, average turnaround time to increase, and the overall throughput to decrease as a result.

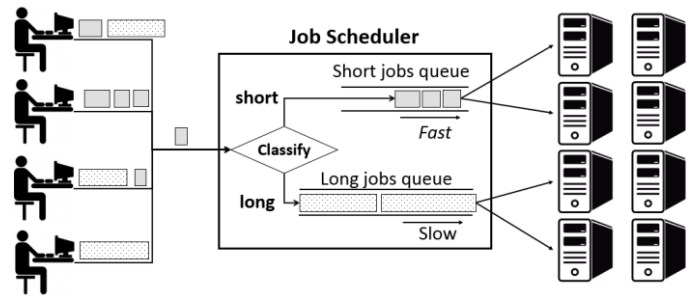

Fig. 1. Separating the short jobs from the long reduces the likelihood that short jobs will be delayed after long ones and improves system performance.

Motivated by the later usage model (server farms), we developed a method that allows predicting the runtime classes of the jobs with high accuracy. Our method is based on applying a log transformation on the runtimes of the jobs (historical records), revealing a mixture of two overlapping Gaussian distributions that represent the short and long jobs. We use the mixture model to determine the distribution parameters and to set the initial separation threshold between the short and long runtime populations.

A key design point for our method is to be able to predict the classes with high accuracy. In order to achieve this, we do not determine the threshold that separates the short jobs from the long in advance (which can lead to an eventual high misclassification rate). Instead, we determine it as part of the evaluation of the classifier: we use a subset of the data that is close to the means of the distributions to train the classifier, and then use the full dataset to select the threshold that optimizes a desired target function.

For class prediction for newly incoming jobs, we use the CART classifier [18], which is suitable for binary classification and can account for both continuous and categorical classifying variables. CART uses a tree optimizing algorithm that minimizes classification error while reducing over fitting by branch pruning.

We applied our method on a job trace obtained from one of Intel's data center installations, and which contained more than one million job records. Setting the target on achieving the best trade-off between misclassifications of short jobs and misclassifications of long jobs resulted in prediction accuracy of 90% (total misclassification rate of 10%) on the independent validation set. The predictions were based on estimated distribution means of 140 and 3,500 seconds for the "short" and "long" classes, respectively, and a separating threshold of 608 seconds.

This paper is organized as follows. Section 2 describes the data we used to train, test and validate our model. Section 3 describes the initial class labeling based on the mixture model analysis. Section 4 reviews the CART model. Section 5 describes the learning algorithm along with the optimal threshold determination procedure for best accuracy. Section 6 describes the results of the study. Section 7 surveys related work and Section 8 concludes the paper.

## II. THE DATA

Our data is based on two traces obtained from one of Intel's data centers installations. The first trace, which was used to train the model, contained around one million records of job that executed in the data center during a period of ten consecutive days. The second trace, which was used to validate the model, contained additional 755,000 records (approximately) of jobs that executed during a period of seven consecutive days. The validation on independent data is important for establishing the robustness of the model obtained in the training stage.

Each record in the traces contained 13 fields pertaining to a particular job. These included three continuous variables: "Submittime", "Starttime" and "Finishtime", indicating when the job was submitted, when it started, and when it finished executing, respectively. A discrete variable, "Iterations", recorded the number of attempts elapsed until the job completed successfully. It ranged from 1 to 764, with a mean of 1.6 and standard deviation of 5.3 (the vast majority of the jobs had up to 20 iterations). In order not to reveal any information about the workload, the traces did not contain any descriptive information about the jobs. Instead, the values in the fields were transformed into discrete values (categorical variables) that can be used for the analysis. In addition, the names were also transformed in order not to reveal infor-mation about the possible meanings of the values.

Table 1 groups the 9 categorical variables and roughly explains the meaning of each group. Table 2 outlines basic statistics on each of the variables.

TABLE I. ROUGH GROUPING OF THE 9 CATEGORICAL VARIABLES

| Group | # of variables | Relates to | Example |
|---|---|---|---|
| A | 3 | Scheduling information | Resources requested by the job |
| B | 2 | Execution-specific information | Command line and arguments |
| C | 4 | Association information | Project and component |

TABLE II. STATISTICS REGARDING THE CATEGORICAL VARIABLES

| Variable | # of categories | # of missing (in training data) |
|---|---|---|
| A1 | 9 | 0 |
| A2 | 7 | 0 |
| A3 | 5 | 0 |
| B1 | 44 | 173 |
| B2 | 22 | 184 |
| C1 | 2 | 0 |
| C2 | 5 | 239 |
| C3 | 6 | 184 |
| C4 | 32 | 0 |

In addition to the above, we defined two additional categorical variables, day and hour, based on the three continuous variables in the trace. These variables indicate the day of the week (1 for Sunday to 7 for Saturday) and the hour of the day (0 to 23), the job was submitted, started, and finished executing, respectively. Figure 2 shows the distribution of the respective temporal categorical variables along the timeline axis. As can be seen, during weekdays longer jobs are typically submitted during the morning hours, with occasional peaks in runtime during evening hours. During weekends, peaks in runtime also occur during the afternoon and evening hours.

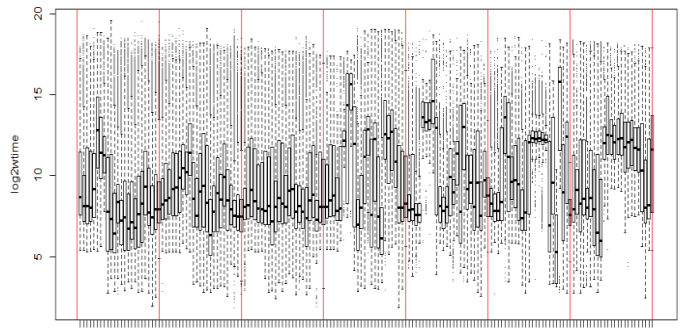

Fig. 2. Runtime boxplots (in log base 2 scale) along time (Sunday through Saturday). The days are separated by vertical red lines. Each tick mark along the time axis marks an hour of the day.

Figure 3 shows the distribution of the jobs runtime "as-is", and after applying a log base 2 transformation on the runtime. As can be seen in Figure 3a, the vast majority of the jobs are short (the shortest job ran 3.5 seconds), and there are few long ones (the longest job ran for nearly 9 days). This well corresponds to previous observations made on the runtime, describing a phenomenon that characterizes many production workloads [11].

Transforming the runtime to the log scale (Figure 3b) reveals a mixture pattern of two main Gaussian-like distributions (sometimes referred to as a "hyper lognormal distribution), with a stretching right tail.

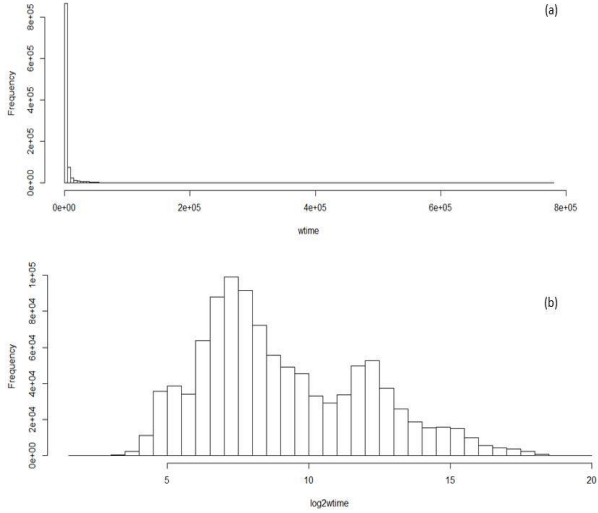

Fig. 3. Histogram of runtime. (a) Raw data (b) After log base 2 transformation.

## III. Class construction by Mixture distribution analysis

The first component in our analysis sets the base for defining the two runtime classes by estimating the mixture distribution parameters. Using this initial step, we can label each job as "short" or "long". Once the predictor variables are selected through a training-and-testing algorithm (see Section 5.1), the model is optimized by selecting the mixture threshold which provides the best performance, namely minimizes the prediction error or approaches the desired sensitivity-specificity combination (see Section 5.2).

The Gaussian (normal) mixture model has the form

$$f(x) = \sum_{m=1}^{M} \alpha_m \phi(x; \mu_m, \Sigma_m)$$

with mixing proportions $\alpha_m$, $\sum_m \alpha_m = 1$, and each Gaussian density has a mean $\mu_m$ and covariance matrix $\Sigma_m$. The parameters are usually fit by the maximum likelihood approach using the EM algorithm, which is a popular tool for simplifying difficult maximum likelihood problems.

Based on the mixture observed in Figure 2, we may define from one to four mixture components. However, for the purpose of this study we decided to focus only on two classes, namely "short jobs" and "long jobs". Thus we model the runtime $Y$ as a mixture of the two normal variables

$$Y_1 \sim N(\mu_1, \sigma_1^2), \qquad Y_2 \sim N(\mu_2, \sigma_2^2).$$

$Y$ can be defined by

$$Y = (1 - \Delta) \cdot Y_1 + \Delta \cdot Y_2,$$

where $\Delta \in \{0, 1\}$ with $\mathbb{P}(\Delta = 1) = \pi$. This generative representation is explicit: generate a $\Delta \in \{0, 1\}$ with probability $\pi$, and then depending on the outcome, deliver either $Y_1$ or $Y_2$. Let $\phi_\theta(x)$ denote the normal density with parameters $\theta = (\mu, \sigma^2)$. Then the density of $Y$ is

$$g_Y(y) = (1 - \pi)\phi_{\theta_1}(y) + \pi\phi_{\theta_2}(y).$$

Suppose we wish to fit this model to our data by maximum likelihood. The parameters are

$$\theta = (\pi, \theta_1, \theta_2) = (\pi, \mu_1, \sigma_1^2, \mu_2, \sigma_2^2).$$

The log-likelihood based on $N$ training cases is

$$l(\theta; Z) = \sum_{i=1}^{N} \log[(1 - \pi)\phi_{\theta_1}(y_i) + \pi\phi_{\theta_2}(y_i)].$$

Direct maximization of $l(\theta; Z)$ is quite difficult numerically due to the sum of terms inside the logarithm. There is, however, a simpler approach. We consider unobserved latent variables $\Delta_i$ taking values 0 or 1 as earlier: if $\Delta_i = 1$ then $Y_i$ comes from distribution 2, otherwise it comes from distribution 1. Suppose we knew the values of the $\Delta_i$'s. Then the log-likelihood would be

$$l(\theta; Z, \Delta) = \sum_{i=1}^{N} [(1 - \Delta_i) \log \phi_{\theta_1}(y_i) + \Delta_i \log \phi_{\theta_2}(y_i)]$$
$$+ \sum_{i=1}^{N} [(1 - \Delta_i) \log \pi + \Delta_i \log(1 - \pi)]$$

and the maximum likelihood estimates of $\mu_1$ and $\sigma_1^2$ would be the sample mean and the sample variance of the observations with $\Delta_i = 0$. Similarly, the estimates for $\mu_2$ and $\sigma_2^2$ would be the sample mean and the sample variance of the observations with $\Delta_i = 1$.

Since the $\Delta_i$ values are actually unknown, we proceed in an iterative fashion, substituting for each $\Delta_i$ in the previous equation its expected value

$$\gamma_i(\theta) = \mathbb{E}(\Delta_i | \theta, Z) = \mathbb{P}(\Delta_i = 1 | \theta, Z),$$

which is also called the responsibility of model 2 for observation $i$.

We use the following procedure, known as the EM algorithm, for the two-component Gaussian mixture:

1. Take initial guesses for the parameters $\hat{\pi}, \hat{\mu}_1, \hat{\sigma}_1^2, \hat{\mu}_2, \hat{\sigma}_2^2$ (see below).

2. Expectation step: compute the responsibilities

$$\hat{\gamma}_i = \frac{\hat{\pi}\phi_{\hat{\theta}_2}(y_i)}{(1-\hat{\pi})\phi_{\hat{\theta}_1}(y_i) + \hat{\pi}\phi_{\hat{\theta}_2}(y_i)}, i = 1, 2, \ldots, N. \quad (1)$$

3. Maximization step: compute the weighted means and variances,

$$\hat{\mu}_1 = \frac{\sum_{i=1}^{N}(1 - \hat{\gamma}_i)y_i}{\sum_{i=1}^{N}(1 - \hat{\gamma}_i)}, \qquad \hat{\sigma}_1^2 = \frac{\sum_{i=1}^{N}(1 - \hat{\gamma}_i)(y_i - \hat{\mu}_1)^2}{\sum_{i=1}^{N}(1 - \hat{\gamma}_i)},$$

$$\hat{\mu}_2 = \frac{\sum_{i=1}^{N} \hat{\gamma}_i y_i}{\sum_{i=1}^{N} \hat{\gamma}_i}, \qquad \hat{\sigma}_2^2 = \frac{\sum_{i=1}^{N} \hat{\gamma}_i (y_i - \hat{\mu}_1)^2}{\sum_{i=1}^{N} \hat{\gamma}_i},$$

and the mixing probability,

$$\hat{\pi} = \frac{\sum_{i=1}^{N} \hat{\gamma}_i}{N}.$$

4. Iterate steps 2 and 3 until convergence.

In the expectation step, we do a "soft" assignment of each observation to each model: the current estimates of the parameters are used to assign responsibilities according to the relative density of the training points under each model. In the maximization step, these responsibilities are used within weighted maximum-likelihood fits to update the estimates of the parameters.

A simple choice for initial guesses for $\hat{\mu}_1$ and $\hat{\mu}_2$ is two randomly selected observations $y_i$. The overall sample variance $\sum_{i=1}^{N} \frac{(y_i - \bar{y})^2}{N}$ can be used as an initial guess for both $\hat{\sigma}_1^2$ and $\hat{\sigma}_2^2$. The initial mixing proportion $\hat{\pi}$ can be set to 0.5. The "mixtools" R package [15, 16] was used for the mixture analysis, with the function "normalmixEM" for parameter and posterior probability (responsibility) estimation.

## IV. THE CART MODEL

The CART (Classification and Regression Trees) model, also named the decision tree model, is an approach for making either quantitative or class prediction. It is non-parametric, thus no assumptions are made regarding the underlying distribution of the predictor variables, enabling CART to handle numerical data that are skewed or multi-modal. It can consider both continuous and categorical predictors, including ordinal ones

CART identifies classifying, or "splitting", variables based on an exhaustive search of all classifying possibilities with the available variables. Useful CART trees can be generated even when there are missing values for some variables, by using "surrogate" variables, which contain information similar to the missing variables.

CART analysis consists of the following steps:

• Tree building, during which a tree is built using recursive splitting of nodes. This process stops when a maximal tree has been produced. The higher the splitter variable in the tree, the higher its importance in the prediction process.

• Tree "pruning", which is a simplification of the tree by cutting nodes off from the maximal tree.

• Optimal tree selection, which selects one tree from the set of pruned trees with the least evidence of over fit.

Each path from the root of a decision tree to one of its leaves can be transformed into a rule. Less complex decision trees are preferred, since they are easier for interpretation and may be more accurate.

## V. MODEL LEARNING AND OPTIMIZATION PROCEDURE

Once labeled data is obtained, a supervised learning technique is used for the purpose of generating a classification rule. In the first stage we select a set of variables that will be included in the model, while evaluating the performance of each model, and in the second stage we obtain a final model by using the selected variables on the full dataset and evaluate it based on the performance target function.

### A. Variable selection

This stage of the analysis is done on a subset of the training data (containing the ten days period), which is extracted as follows. Since the two observed runtime distributions overlap, we select observations that are within 0.5 standard deviations off the two means, such that they will be distant from the overlapping region and will belong to the corresponding classes with high certainty (Figure 4). A total of 257,467 obser-vations labeled short=1 (belonging to the Gaussian population with the smaller mean) and 192,205 observations with short=0 (belonging to the Gaussian population with the larger mean) were selected. Together they made around 43% of the data.

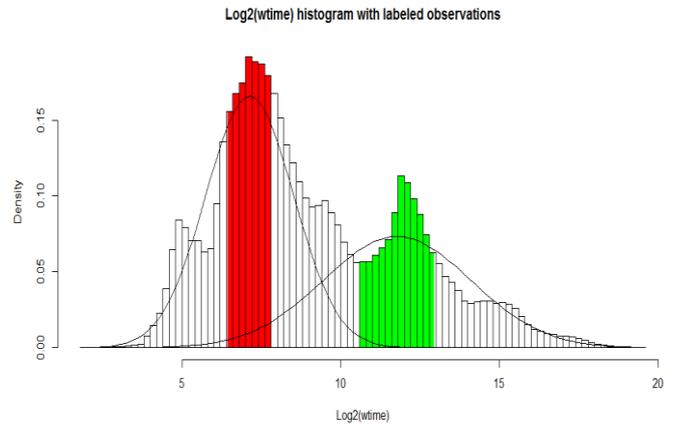

Fig. 4. Runtime (in log base 2 scale) density. The red and green colored regions mark the observations selected for the learning process.

We then perform a five-fold cross-validation procedure in order to select variables and evaluate each model, by iteratively dividing our data in random into a training set (80% of the learning data) and an evaluation set (20% of the learning data) and implementing the CART model with the mixture threshold of 0.5. The importance measure, which considers how high the splitting variable is in the tree, is averaged across all iterations for each variable, and we select the variable having an importance score above the baseline level.

Once a set of variables is selected for each model type, a model is fit to the full training data. At this stage, we account for the two types of misclassification error, the false "positive" classification and the false "negative" classification. Defining classification into "short" as "positive", the former refers to the erroneous classification of a long job into the "short" class, and the latter refers to the erroneous classification of a short job into the "long" class.

In the job runtime context, sensitivity is defined as the proportion of short jobs classified as short, while the specificity is defined as the proportion of long jobs classified as long. Subtracting the specificity from 1 will give the proportion of long jobs erroneously classified as short. For the CART classifier, we choose the mixture threshold that yields the best tradeoff between these two errors. The full set of sensitivity-specificity combinations can be summarized in a pseudo-ROC (Receiver Operating Characteristic) curve, in which the sensitivity is plotted against 1-specificity, for each threshold of the probability obtained in the mixture model (the final value obtained for equation 1).

A model performing a perfect discrimination has an ROC curve that passes through the upper left corner (100% sensitivity, 100% specificity). Therefore the closer the ROC curve to the upper left corner, the higher the overall accuracy of the test. Yet, the consequences, or costs, of each type of error may vary among applications and among policy makers. Thus the optimal threshold may allow higher weight to one of the errors on account of the other.

## VI. RESULTS

### A. Class definition

Implementing the mixture model clustering approach, two Gaussian families underlying the runtime distribution (on the log base 2 scale) were defined. The density estimates are super imposed on the runtime density in Figure 5. The parameters for each family and the mixing proportions are detailed in Table 3.

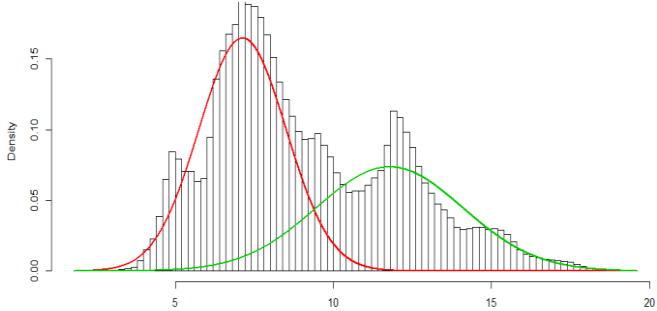

Fig. 5. Density estimates obtained by mixture analysis for the two families underlying the runtime distribution (on the log base 2 scale). The red line marks the estimated density for the "short" class, while the green line marks the estimated density for the "long" class.

The mixture analysis also yields the posterior probability, as defined by equation (1), for each observation to belong to the "short" class. For a probability threshold of 0.5, 631,059 observations (nearly 60%) are classified into the "short" class, while 411,053 observations are classified into the "long" class. Once we find a classifier (see the next subsection), we refine this threshold to optimize the sensitivity-specificity tradeoff.

TABLE III. DENSITY PARAMETER ESTIMATES OBTAINED BY THE MIXTURE ANALYSIS

|  | First Gaussian Family | | Second Gaussian Family | |
|---|---|---|---|---|
| **Mixing Proportion** | 0.57 | | 0.43 | |
| **Mean** | 7.13 | $2^{7.13}$ | 11.78 | $2^{11.78}$ |
| **Standard Deviation** | 1.38 | $2^{1.38} = 2.60\ sec$ | 2.32 | $2^{2.32} = 4.99\ sec$ |

### B. Classifying by CART

Applying the CART classifier on the training data through the cross-validation procedure, six variables obtained high importance scores (Figure 6). The classifier achieved a total misclassification error of 3.5%.

A model containing the six selected variables was fit to the full dataset for a series of class mixture threshold. The pseudo-ROC in Figure 7 presents the sensitivity-specificity combinations obtained for a set of threshold probability used for the mixture distribution. The best performing model is the one using the threshold of 0.45, which achieves sensitivity of 92.5% and specificity of 91.1%, with a total misclassification error of 8.9%. This threshold corresponds to a runtime of 9.25 on the log scale, or 608 seconds on the original scale.

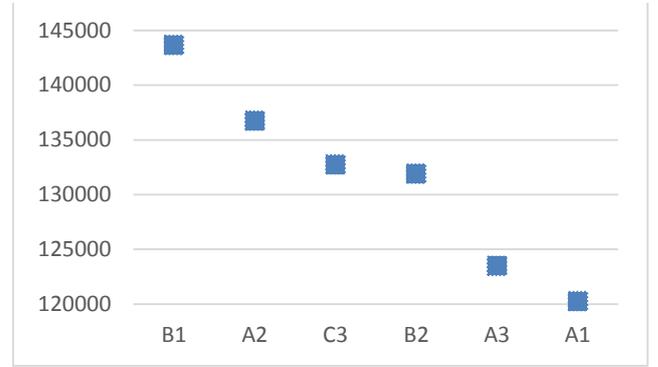

Fig. 6. Top ranking importance scores for the CART model, averaged across 150 cross-validation iterations.

The selected model yielded a tree containing four of the six variables that were tried (Figure 8). The total misclassification rate was 8.08%. Implementing the obtained tree on the validation data resulted in a total misclassification error of 9.17%, with specificity of 91.5% and sensitivity slightly beyond 90%.

## VII. RELATED WORK

Supplying the scheduler with information on how long the jobs are expected to run has always been a challenging task. In general, two approaches were used for this purpose. The first is to ask the users to supply the information, and the other is to try and predict the runtimes automatically using historical data on jobs that have already completed.

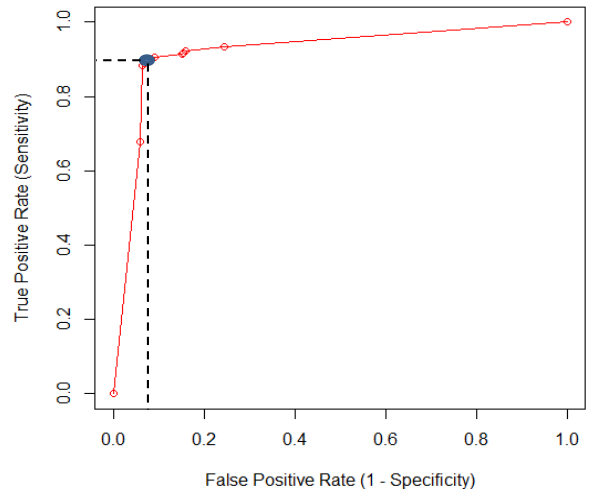

Fig. 7. The pseudo-ROC curve obtained by the CART classifier for the full training data. The blue circle marks the optimal tradeoff between sensitivity and specificity (enhanced by the dashed lines), obtained for mixture probability threshold of 0.45.

Asking the users to estimate the runtimes has been shown to be highly inaccurate, as users tend of overestimate the runtimes in order to prevent the scheduler from killing their jobs [1]. Furthermore, Tsafrir et al. [2] has observed that the users further tend to "round" the estimates, thereby limiting the scheduler's ability to optimize the schedule. Bailey et al. [7] have shown that users are quite confident of their estimates, and that most likely they will not be able to improve it much.

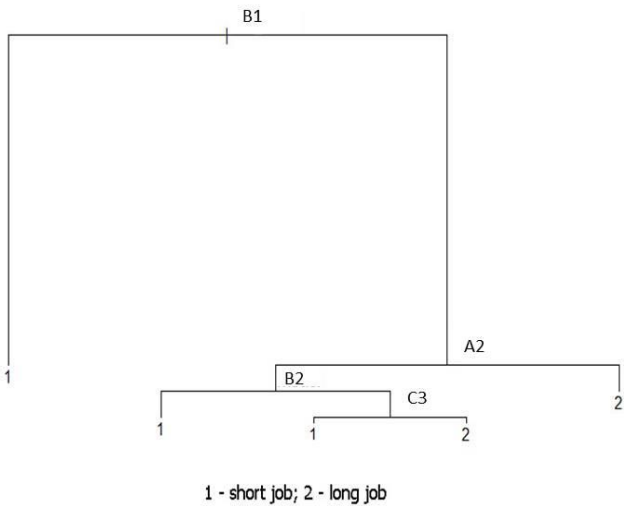

Fig. 8. The final tree obtained on the full data.

Predicting the runtimes automatically is therefore the default alternative. This is usually composed of two steps:

1. Identifying classes of similar jobs within the historical jobs records, and
2. Using the aforementioned classes to predict the runtimes for newly submitted jobs.

Gibbons [4] and Downey [5] classified the jobs using a statically defined set of attributes, e.g. user, executable, queue, etc. For newly submitted jobs, Gibbons used the 95th percentile of the runtimes in the respective class, while Downey used a statistical model that was based on a log-uniform distribution of the runtimes in order to provide the prediction.

Smith et al. [6] suggested the use of genetic algorithms to refine the selection of at-tributes used for the classification, and showed that this can yield up to 60% improvement in accuracy compared to the static approaches. Respectively, Kapadia et al. [8] used instance-based learning and Krishnaswamy et al. [9] applied rough-set theory. Finally, Tsafrir et al. [10] showed that complicated prediction techniques may not be required if the scheduling algorithm itself can be modified, and suggested to average the last two jobs by the same user in the history.

These techniques, however, were mainly designed for parallel systems, in which the scheduler needs to the actual runtimes of the jobs (use numeric predictions) to be able to optimize the schedule. For the server farm usage model which we target in our work, sophisticated scheduling algorithms are not required, and it was shown that it is enough to simply separate the jobs into short and long to improve performance [12]. Our work provides the facility to enable this separation and hence forms the basis for enabling such systems.

## VIII. CONCLUSIONS

Predicting the runtimes of jobs using actual numeric values is of high importance for parallel systems. Here, fragmentation is a big concern, and in order to minimize it (namely to fill the holes in the schedule), it is important for the scheduler to know the exact runtime of the jobs. For other types of systems like server farms used for software testing, it is enough to only predict the runtime classes of the jobs e.g., short or long, in order to send them to the right queue and improve performance.

Motivated by the later usage model, we developed a method that facilitates highly accurate prediction of the job runtime class. Our method leverages the fact that the runtimes may be represented as a mixture of two or more distributions, in order to train a classifier that will be used to predict the runtime classes of newly incoming jobs. In order to achieve high accuracy, the threshold that separates the short jobs from the long is determined during the evaluation of the classifier. In a real system this threshold can be periodically communicated to the scheduler to help it decide on the right allocation of resources for the different job classes.

Our work is based on a single data set that we obtained, and included validation on data independent of the training data. Yet, in spite of its promising results, additional testing is required on more data sets in order to establish complete confidence in the robustness of our method. However, due to the size of our data (over one million jobs), and the fact that the mixture distribution is known to be evident in the many real-world workloads, we have a strong reason to believe that with small adjustments e.g., to the number of classes, our method can be tuned to sustain those workloads as well. Obtaining and experimenting with more data sets is part of our future research agenda.

ACKNOWLEDGMENT

We thank Prof. Dror G. Feitelson of the Hebrew University, Israel, for the useful comments, Evgeni Korchatov from Intel for helping to obtain and prepare the data and Eran Smadar from Intel for supporting this work.